\documentclass[10pt,twocolumn,letterpaper]{article}

\usepackage[pagenumbers]{cvpr} 

\definecolor{cvprblue}{rgb}{0.21,0.49,0.74}
\usepackage[pagebackref,breaklinks,colorlinks,allcolors=cvprblue]{hyperref}

\usepackage{overpic}

\title{UniModel: A Visual-Only Framework for Unified \\ Multimodal Understanding and Generation}

\author{
Chi Zhang
\quad
Jiepeng Wang
\quad
Youming Wang
\quad
Yuanzhi Liang
\quad
Xiaoyan Yang\\
Zuoxin Li
\quad
Haibin Huang
\quad
Xuelong Li
 \\
{\normalsize Institute of Artificial Intelligence (TeleAI), China Telecom} 
}

\begin{document}
\maketitle
\begin{abstract}
We present UniModel, a unified generative model that jointly supports visual understanding and visual generation within a single pixel-to-pixel diffusion framework. Our goal is to achieve unification along three axes: the model, the tasks, and the representations.

At the representation level, we eliminate modality discrepancies by mapping both text and images into a shared visual space: textual prompts are rendered as painted text images on a clean canvas, and all inputs and outputs are treated purely as RGB pixels. This yields a fully vision-native formulation of multimodal learning.

At the task level, a broad range of vision–language problems are cast as pixel-to-pixel transformations in this visual space. For understanding tasks, the model takes an RGB image and produces a painted text image that visually encodes the semantic prediction. For generation tasks, painted text images serve as visual conditions that guide realistic and semantically aligned image synthesis. Captioning and text-to-image generation thus become different directions of the same underlying visual translation process.

At the model level, we instantiate a single Unified Diffusion Transformer trained with rectified flow in pixel space. A shared backbone jointly learns bidirectional mappings between natural images and painted text images, with lightweight task embeddings to specify the desired direction. Training is entirely pixel-based, avoiding mismatched objectives across tasks and enabling direct cross-modal supervision—for example, editing a word in the painted caption naturally steers the corresponding image region.

Experiments on text-to-image synthesis and image-to-text understanding demonstrate strong cross-modal alignment and emergent controllability such as cycle-consistent image–caption–image loops. Our initial exploration suggests that unifying model, tasks, and representations in a single visual space is a promising paradigm for general-purpose multimodal intelligence.
\end{abstract}    
\section{Introduction}
\label{sec:intro}

Recent advances in multimodal AI have been driven by two parallel lines of progress: large language models (LLMs) that excel at text understanding and generation \cite{brown2020language, touvron2023llama}, and diffusion-based vision models that produce high-fidelity images and videos  \cite{ho2020denoising, rombach2022high}. These developments have motivated a surge of “unified” architectures that aim to handle both vision and language within a single framework, including BLIP-style models \cite{li2022blipbootstrappinglanguageimagepretraining,li2023blip2bootstrappinglanguageimagepretraining,dai2023instructblipgeneralpurposevisionlanguagemodels}, Flamingo \cite{alayrac2022flamingovisuallanguagemodel}, LLaVA variants \cite{liu2023visualinstructiontuning,liu2024improvedbaselinesvisualinstruction}, and more recent autoregressive and diffusion-based backbones such as Kosmos-2 \cite{peng2023kosmos2groundingmultimodallarge}, Emu \cite{sun2024emugenerativepretrainingmultimodality,wang2024emu3nexttokenpredictionneed,cui2025emu35nativemultimodalmodels} , and VideoPoet \cite{kondratyuk2024videopoetlargelanguagemodel}.

However, despite using a shared backbone, most of these systems remain only partially unified. They typically share parameters at the model level, but still rely on heterogeneous representations and fragmented task formulations. Images and text are encoded by separate modules into different embedding spaces; understanding and generation are optimized with distinct objectives and decoders; and supervision signals are mismatched across tasks. As a result, these models often treat “image → text” and “text → image” as fundamentally different problems, which can lead to conflicting gradients, suboptimal cross-modal alignment, and limited bidirectional control.

We argue that backbone sharing alone is not enough. For truly general-purpose multimodal intelligence, a unified model should operate coherently along three axes:

(1) Representation-level unification: different modalities should be expressed in a compatible form, so that the model does not need to constantly bridge gaps between discrete symbols and continuous pixels.

(2) Task-level unification: diverse multimodal tasks should reduce to a small set of common primitives, enabling shared learning signals and consistent objectives.

(3) Model-level unification: a single parameterization should be able to support these tasks in both directions, rather than relying on separate heads or decoders for understanding vs. generation.

This perspective leads to a simple question: instead of aligning text and images across heterogeneous spaces, can we homogenize their representation so that they can be processed as the same kind of signal?

\begin{figure}
    \centering
    \begin{overpic}[width=\linewidth]{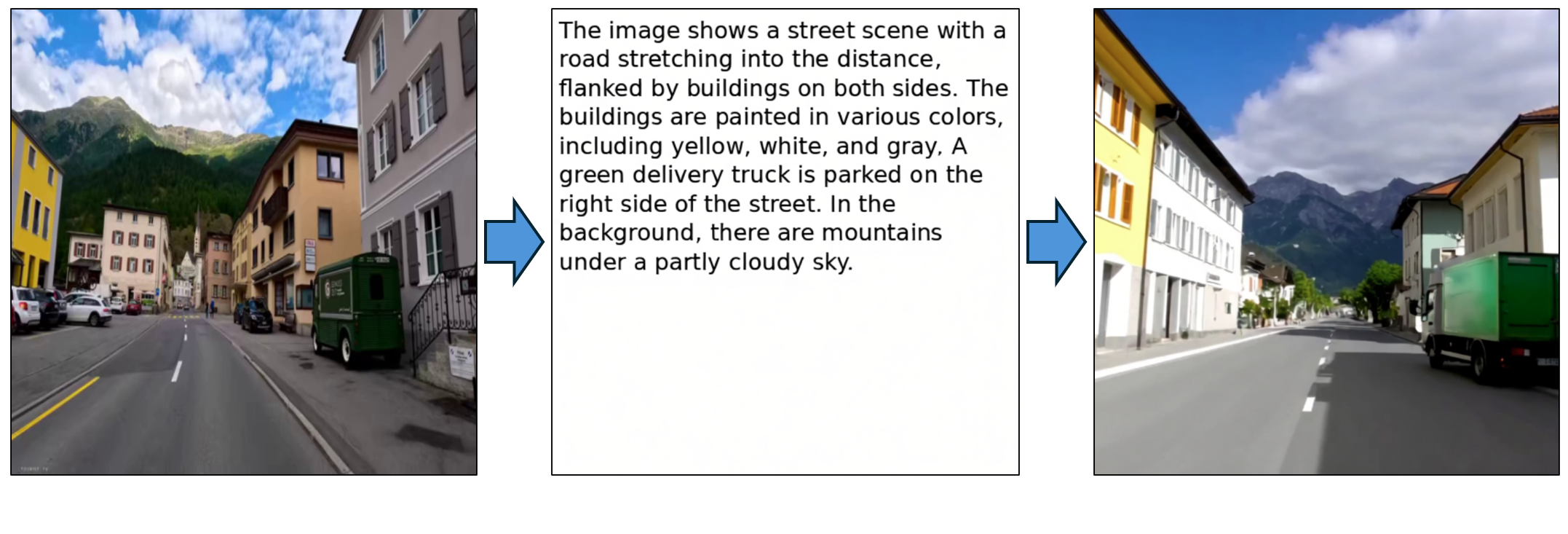}
        \put(18, 0){\small (a) Image-to-text}
        \put(58, 0){\small (b) Text-to-image}
    \end{overpic}
    \caption{Unified visual representation. (a) Given an RGB image, the model outputs its corresponding painted text image. (b) Using the painted text image as a condition, the model generates a realistic RGB image.}
    \label{fig:placeholder}
\end{figure}

In this work, we advocate a fully visual view of multimodal learning. We map both text and images into a shared visual space by rendering textual content as painted text images on a blank canvas. This approach builds on a growing body of work that treats text as a visual signal for tasks like language modeling \cite{rust2023languagemodellingpixels}, long-context compression \cite{wei2025deepseekocrcontextsopticalcompression, lu2024textpixeladvancinglongcontext}, and generative conditioning \cite{li2023glyphdiffusiontextgenerationimage, yang2023glyphcontrolglyphconditionalcontrol}. Once text is represented as an RGB image, captions, questions, and answers become visual objects, just like natural images. This yields representation-level unification: all inputs and outputs are RGB images, and the model operates entirely in a visual domain.
Under this representation, a wide range of vision–language tasks collapse into the same primitive: pixel-to-pixel translation. Visual understanding tasks, such as image captioning or visual question answering, can be cast as RGB-image → painted-text-image transformations. Visual generation tasks, such as text-to-image synthesis, become painted-text-image → RGB-image transformations. From this viewpoint, understanding and generation are no longer separate pipelines, but opposite directions of the same visual mapping, achieving task-level unification.

Building on this idea, we introduce UniModel, a unified diffusion framework that treats multimodal understanding and generation as a single pixel-to-pixel problem. At the model level, UniModel is instantiated as a bidirectional diffusion transformer \cite{peebles2023scalable} trained with a rectified-flow \cite{liu2022flowstraightfastlearning} objective on VAE latents \cite{rombach2022high}. A shared backbone jointly learns all mappings between natural and painted images, while lightweight task embeddings indicate the desired direction (e.g., understanding vs. generation). Training is performed entirely in pixel space, with consistent diffusion objectives for both directions, achieving model-level unification without task-specific decoders or losses.

This three-fold unification has several practical benefits. First, it provides a clean and symmetric formulation of image$\leftrightarrow$text relations, which encourages stronger cross-modal alignment and cycle consistency. Second, it enables natural forms of controllability: editing a word in the painted caption becomes a local visual edit that can steer the corresponding region of the generated image. Third, it simplifies the overall system design, avoiding hand-crafted interfaces between separate language and vision modules.

In summary, this work makes the following contributions:
\begin{itemize}
    \item We propose a representation-level unification strategy that maps both text and images into a shared visual space via painted text images, enabling a fully vision-native formulation of multimodal learning.

   \item We cast a broad range of vision–language tasks as pixel-to-pixel translations between natural and painted images, achieving task-level unification of visual understanding and generation.

   \item We introduce UniModel, a single bidirectional diffusion transformer trained with rectified-flow objectives on VAE latents, which realizes model-level unification with shared parameters and lightweight task embeddings.

   \item We demonstrate on text-to-image synthesis and image-to-text understanding that UniModel attains competitive performance, strong cross-modal alignment, and emergent controllability such as cycle-consistent image–caption–image reconstruction.

\end{itemize}

\section{Related Work}
\label{sec:formatting}

\paragraph{Unified models}

The pursuit of foundation models that holistically unify visual and textual understanding and generation has catalyzed the exploration of several architectural paradigms. A distinct approach leverages unified discrete diffusion frameworks, which model multimodal data in a non-autoregressive manner as seen in
\cite{shi2025mudditliberatinggenerationtexttoimage, wang2025fudokidiscreteflowbasedunified, yang2025mmadamultimodallargediffusion, swerdlow2025unifiedmultimodaldiscretediffusion, li2025dualdiffusionunifiedimage}. However, the predominant approach centers on extending Multimodal Large Language Models (MLLMs) with autoregressive (AR) generation capabilities, differentiated by their visual encoding strategies. One strategy involves Pixel Encoding, where images are represented by discrete visual tokens, forming a visual vocabulary for the AR model. This is exemplified by works such as
\cite{wang2025selftokdiscretevisualtokens, lin2025toklipmarryvisualtokens, wu2025harmonizingvisualrepresentationsunified, tang2025ugenunifiedautoregressivemultimodal, li2024synergenvlsynergisticimageunderstanding, wu2025liquidlanguagemodelsscalable, kou2025orthusautoregressiveinterleavedimagetext, yang2025mmarlosslessmultimodalautoregressive, wang2024emu3nexttokenpredictionneed, cui2025emu35nativemultimodalmodels,chern2024anoleopenautoregressivenative, chameleonteam2025chameleonmixedmodalearlyfusionfoundation,liu2025worldmodelmillionlengthvideo}. A more prevalent strategy is Semantic Encoding, which utilizes semantically rich tokens derived from powerful vision encoders like CLIP. This paradigm is foundational to a vast array of models including
 \cite{lin2025bifrost1bridgingmultimodalllms, geng2025xomnireinforcementlearningmakes, wang2025ovisu1technicalreport, chen2025unicode2cascadedlargescalecodebooks, wu2025omnigen2explorationadvancedmultimodal, han2025visiondialectunifyingvisual, li2025uniforkexploringmodalityalignment, lin2025uniworld, xu2025piscesautoregressivefoundationmodel, song2025dualtokenunifyingvisualunderstanding, ma2025unitokunifiedtokenizervisual, zhao2025qliptextalignedvisualtokenization, tong2024metamorphmultimodalunderstandinggeneration, wang2024illumeilluminatingllmssee, fang2024pumaempoweringunifiedmllm, wu2025vilauunifiedfoundationmodel, li2024minigeminiminingpotentialmultimodality, tian2024mminterleavedinterleavedimagetextgenerative, zhu2023vlgptgenerativepretrainedtransformer, sun2024generativemultimodalmodelsincontext, dong2024dreamllmsynergisticmultimodalcomprehension, jin2024unifiedlanguagevisionpretrainingllm, sun2024emugenerativepretrainingmultimodality}. Further refinement is achieved through Learnable Query Encoding, where a set of learnable vectors distill visual features into fixed-length representations, a technique central to models like
 \cite{xu2025tbacuniimageunifiedunderstanding, tang2025unilipadaptingclip, mingomni2025mingomniaunified, openuni2025openunisimplebaseline, blip3o2025blip3ofamilyfullyopen, mingliteuni2025mingliteuniadvancements, nexusgen2025nexusgenaunifiedmodel, metaqueries2025transfermodalitiesmetaqueries, seedx2024seedxmultimodalmodels, seedllama2023makingllmasee, ge2023plantingseedvisionlarge}. Concurrently, Hybrid Encoding strategies combine semantic and pixel-level features through either pseudo-decoupled
 \cite{janus2024janusdecouplingvisualencoding, januspro2025janusprounifiedmultimodal, mindomni2025mindomniunleashingreasoning, unifluid2025unifiedautoregressivevisual, omnimamba2025omnimambaefficientunified, skywork2025skyworkunipicunifiedautoregressive} or jointly trained
 \cite{vargpt2025vargptunifiedunderstandinggeneration, tokenflow2024tokenflowunifiedimagetokenizer, musevl2024musevlmodelingunifiedvlm, semhitok2025semhitokunifiedimagetokenizer, illumeplus2025illumeplusilluminatingunified, showo22025showo2improvednativeunified, unitoken2025unitokenharmonizingmultimodal}
 visual tokenizers. Bridging these paradigms, a third family of MLLM (AR+Diffusion) models integrates diffusion mechanisms into the AR framework for high-fidelity image decoding, as demonstrated by
 \cite{lmfusion2024lmfusionadaptingpretrained, monoformer2024monoformeronetransformer, showo2024showoonesingletransformer, transfusion2024transfusionpredictnexttoken, bagel2025emergingpropertiesunified, mogao2025mogaoomnifoundationmodel, janusflow2024janusflowharmonizingautoregression}
 with hybrid encoding. The ultimate ambition of this field is manifested in Any-to-Any models, which aim to process and generate arbitrary sequences of multimodal inputs, a frontier being explored by
 \cite{nextgpt2023nextgptanytoanymultimodal, unifiedio22023unifiedio2scalingautoregressive, videolavit2024videolavitunifiedvideolanguage, anygpt2024anygptunifiedmultimodal, xvila2024xvilacrossmodalityalignment, mio2024miofoundationmodelmultimodal, spider2024spideranytomanymultimodal, omnilow2024omniflowanytoanygeneration, m2omni2025m2omniadvancingomnimllm}.

\paragraph{Unified representation via visual tokens}

An emergent paradigm in multimodal understanding seeks to bypass discrete tokenization by processing text as a visual modality. This principle was established by models like
\cite{rust2023languagemodellingpixels}, which render text into images to enhance cross-lingual transfer. This unification is advanced by models like
\cite{tschannen2023clippoimageandlanguageunderstandingpixels}, using a single encoder for images and rendered text, and
\cite{jiang2024trex2genericobjectdetection}, which fuses visual and text prompts. In document AI, models like
\cite{tang2023unifyingvisiontextlayout} process text and layout as a unified visual input, while frameworks like
\cite{zhu2024unitunifyingimagetext} train a single vision encoder for both image and text recognition. This focus on fine-grained text extends to using explicit glyph-level features as visual conditions, as seen in
\cite{yang2023glyphcontrolglyphconditionalcontrol, gillani2025textpixsglyphconditioned}. A novel application is long-context compression, where models like
\cite{lu2024textpixeladvancinglongcontext,wang2024visincontext, wei2025deepseekocrcontextsopticalcompression} compress text into a compact image representation. Collectively, these methods underscore a significant shift towards treating language as a visual signal at the input stage.

\paragraph{Unified generation via visual representations}

The application of diffusion models to discrete text has primarily spurred two research directions: developing diffusion processes directly in the discrete token space
\cite{austin2023structureddenoisingdiffusionmodels,qian2023diffusionglancingtransformerparallel,he2022diffusionbertimprovinggenerativemasked,zhou2023diffusionnatselfpromptingdiscretediffusion}, on word embeddings for sequence-to-sequence tasks
\cite{gong2023diffuseqsequencesequencetext}, and, more prominently, operating in a continuous proxy space by applying the diffusion-denoising process on word embeddings before projecting back to a vocabulary
\cite{hoogeboom2021argmaxflowsmultinomialdiffusion,li2022diffusionlmimprovescontrollabletext,liu2023composabletextcontrolslatent,gong2023diffuseqsequencesequencetext,han2023ssdlmsemiautoregressivesimplexbaseddiffusion,strudel2022selfconditionedembeddingdiffusiontext,dieleman2022continuousdiffusioncategoricaldata,gao2024empoweringdiffusionmodelsembedding,lovelace2023latentdiffusionlanguagegeneration,yuan2023seqdiffuseqtextdiffusionencoderdecoder,lin2023textgenerationdiffusionlanguage,zheng2024reparameterizeddiscretediffusionmodel,ye2024dinoiserdiffusedconditionalsequence,li2023glyphdiffusiontextgenerationimage}. Distinct from these token-based strategies, an emergent paradigm reframes text generation as a task within the visual domain, thereby circumventing the constraints of traditional vocabularies. This principle was pioneered by methods like
\cite{li2023glyphdiffusiontextgenerationimage}, which leverages diffusion models to synthesize a complete glyph image of the target text for subsequent decoding into a symbolic sequence, and
\cite{tai2024pixarautoregressivelanguagemodeling}, which performs direct auto-regressive generation in pixel space. The concept has been extensively applied to not only font generation
\cite{he2023difffontdiffusionmodelrobust,Fu_2024_CVPR,thamizharasan2024vecfusionvectorfontgeneration}, but also the creation of consistent, high-fidelity text effects via shape-adaptive diffusion models like
\cite{mu2024fontstudioshapeadaptivediffusionmodel}, where the creation of individual characters is treated as a conditional image synthesis problem solved via a diffusion process. A higher level of integration is demonstrated by large-scale multimodal models such as
\cite{wu2025qwenimagetechnicalreport,cui2025emu35nativemultimodalmodels}, which achieve native, high-quality rendering of complex text as an integral component of the generated image within a unified framework. Furthermore, this philosophy has been operationalized through external control mechanisms; techniques based on ControlNet
\cite{wang2025reptextrenderingvisualtext,peong2024typographictextgenerationofftheshelf}, utilize visual layouts as guidance to precisely "paint" text within the latent space of an image. Our work is situated at the confluence of these advancements, employing a unified Diffusion Transformer (DiT) architecture that operates directly on pixel representations to seamlessly generate both text (as images) and photorealistic images within a single, parallel model.

\section{Method}

\begin{figure*}
    \centering
    \begin{overpic}[width=\linewidth]{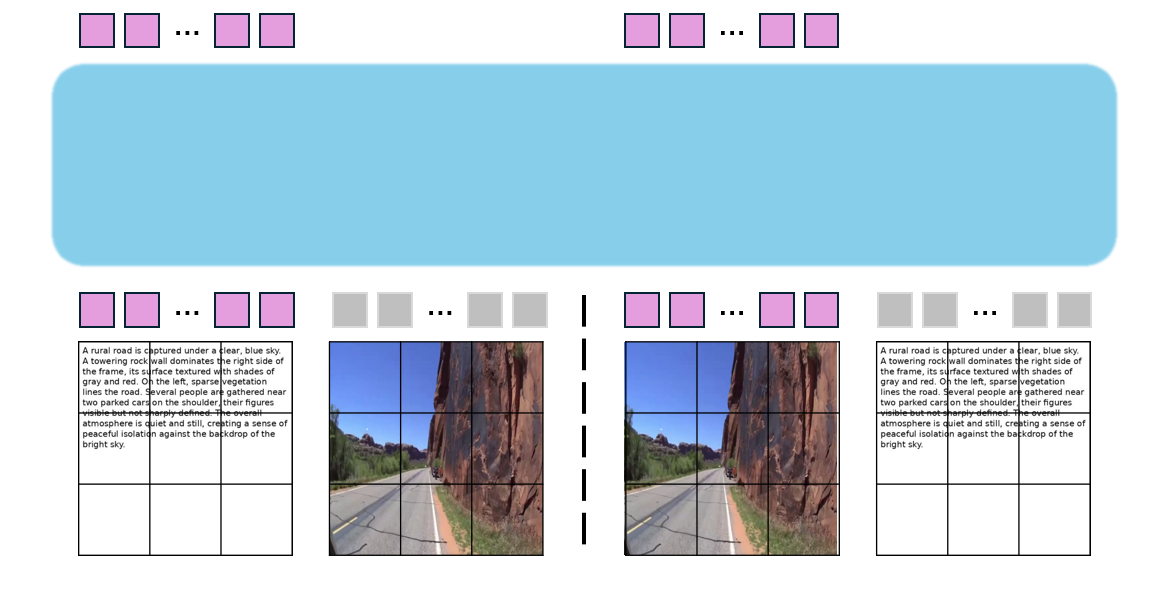}
        \put(18,0){\small (a) Image-to-text}
        \put(68, 0){\small (b) Text-to-image}
        \put(45, 36){\large UniModel}
    \end{overpic}

    \caption{Unified bidirectional generation. (a) Image-to-text: the model encodes an input image into conditional states and generates its text-painted counterpart. (b) Text-to-Image: provided a painted text image, the model synthesizes a realistic RGB image using the same unified architecture. 
    }
    \label{fig:method_overview}
\end{figure*}

In this section, we introduce UniModel, a unified pixel-to-pixel diffusion framework that solves both visual understanding (image~$\rightarrow$~text) and visual generation (text~$\rightarrow$~image) within a single architecture. 
This section is organized as follows: 
Sec.~\ref{sec:unirep} describes the unified visual representation. 
Sec.~\ref{sec:unimodel} introduces the bidirectional diffusion transformer.  
Sec.~\ref{sec:training} presents the unified rectified-flow training objective.  Figure~\ref{fig:method_overview} illustrates the overall pipeline. 

\subsection{Unified Visual Representation}
\label{sec:unirep}

To unify image generation and understanding within a single framework, we express all inputs and outputs—natural images as well as textual prompts—directly in pixel space. This introduces a fundamental representational shift: semantic content, including text, is modeled visually rather than symbolically. Instead of mapping text to discrete token embeddings, we render it as RGB images, enabling learning and inference to occur entirely within a single visual modality.

\paragraph{Text as Pixels.}
We convert textual sequences into RGB images by rendering each prompt, caption, question, or answer onto a blank $512 \times 512 \times 3$ canvas using a fixed font and line spacing, producing a \emph{painted text image}. This 2D rendering preserves the underlying semantics while encoding text in a spatially structured form that is naturally compatible with both convolutional and transformer-based vision modules. For sequences that exceed the available canvas area, we truncate the text to maintain consistent input dimensionality across samples.

\paragraph{A Single Visual Modality.}
Once rendered, painted text images and natural RGB images share the same dimensionality, the same pixel-level representation, and the same preprocessing pipeline. 
Both types of images are encoded through the same VAE to produce latent variables that lie in a unified latent space. As a result, UniModel can treat multimodal tasks as transformations between two visual domains:
\[
\text{RGB image} \;\longleftrightarrow\; \text{painted text image}.
\]

This formulation offers several advantages.  
(1) \textbf{Unified representation}: both text and images are treated uniformly as visual signals, eliminating modality gaps between discrete symbols and continuous pixels.  
(2) \textbf{Unified objective}: understanding and generation share the same rectified-flow diffusion loss, removing the need for cross-entropy decoders, autoregressive language heads, or specialized text objectives.  
(3) \textbf{Unified architecture}: a single diffusion transformer handles all multimodal tasks bidirectionally, enabling strong cross-modal consistency, natural cycle inference, and seamless controllability (e.g., editing a word in the painted caption directly influences corresponding image regions).

\subsection{ Unified Diffusion Transformer} \label{sec:unimodel}
Our model builds on the MMDiT architecture used in Qwen-Image and consists of two components:
(i) a visual encoder module that processes both RGB images and painted text images as model conditions, and
(ii) a diffusion-based generative model that produces either RGB images or rendered text, depending on the task.

All inputs—natural or text-rendered—are encoded by a shared visual encoder into semantic hidden states, which condition the diffusion transformer. A pretrained VAE maps images to a latent space where the rectified-flow diffusion process operates. For understanding tasks, the model takes an RGB image and generates a painted text image; for generation tasks, the input is a painted text image and the output is RGB. This fully visual pipeline removes the need for modality-specific encoders or cross-modal fusion, enabling true architectural unification.

To train the model bidirectionally, we randomly swap input–output pairs (RGB $\rightarrow$ text or text $\rightarrow$ RGB) with equal probability, encouraging symmetric learning of rendering and reconstruction. To further clarify the desired direction, we introduce two learnable task embeddings—one for understanding and one for generation—which are concatenated with conditioning features before diffusion. These embeddings provide explicit task awareness, improving controllability and convergence.

\subsection{Training} \label{sec:training}
\label{sec:training}

To train the model, we adopt \emph{rectified flow}, which improves stability by predicting the instantaneous velocity along the optimal transport path from data to prior. Let $\mathbf{z}_0$ be the VAE latent of the target image (RGB or painted text), and let $\mathbf{z}_1 \sim \mathcal{N}(\mathbf{0}, \mathbf{I})$ denote a Gaussian prior. Rectified flow defines a straight-line interpolation:
\[
\mathbf{z}(t) = (1 - t)\mathbf{z}_0 + t\mathbf{z}_1, \quad t \in [0, 1],
\]
with ground-truth velocity field
\[
\mathbf{v}^\ast(\mathbf{z}(t), t) = \mathbf{z}_1 - \mathbf{z}_0.
\]

Given conditioning features $\mathbf{c}$ from the visual encoder, a task embedding $\mathbf{e}_\text{task}$, and time step $\qquad t \sim \mathcal{U}(0, 1)$, the diffusion transformer $f_\theta$ is trained to match this velocity using the objective
\[
\mathcal{L}_\text{flow}
= \mathbb{E}_{\mathbf{z}_0,\,\mathbf{z}_1,\,t}
\left[
\left\|
\mathbf{v}^\ast(\mathbf{z}(t), t)
-
f_\theta\big(\mathbf{z}(t),\, t,\, \mathbf{c},\, \mathbf{e}_\text{task}\big)
\right\|_2^2
\right]
.
\]

Under bidirectional training, paired RGB and painted-text latents are randomly assigned as source or target (RGB~$\rightarrow$~text or text~$\rightarrow$~RGB), allowing a single objective to supervise both generation and understanding.

\section{Experiments}

\subsection{Implementation details}
We build our Unified Diffusion Transformer upon the MMDiT (Multi-Modal Diffusion Transformer) architecture, inheriting its block design, attention mechanisms, and adaptive layer normalization (AdaLN) for conditional generation.

\subsection{Text to image generation}
For text-to-image generation, our model takes a painted text image as the conditioning input instead of raw text tokens. This design removes the need for separate text encoders or cross-modal fusion modules, allowing the entire generation pipeline to operate purely within the visual latent space. However, because existing unified or diffusion-based models rely on learned textual embeddings—and none of them accept visualized text as input—direct quantitative comparison under the same input paradigm is not feasible.

\begin{figure*}
    \centering
    \begin{overpic}[width=0.9\linewidth]{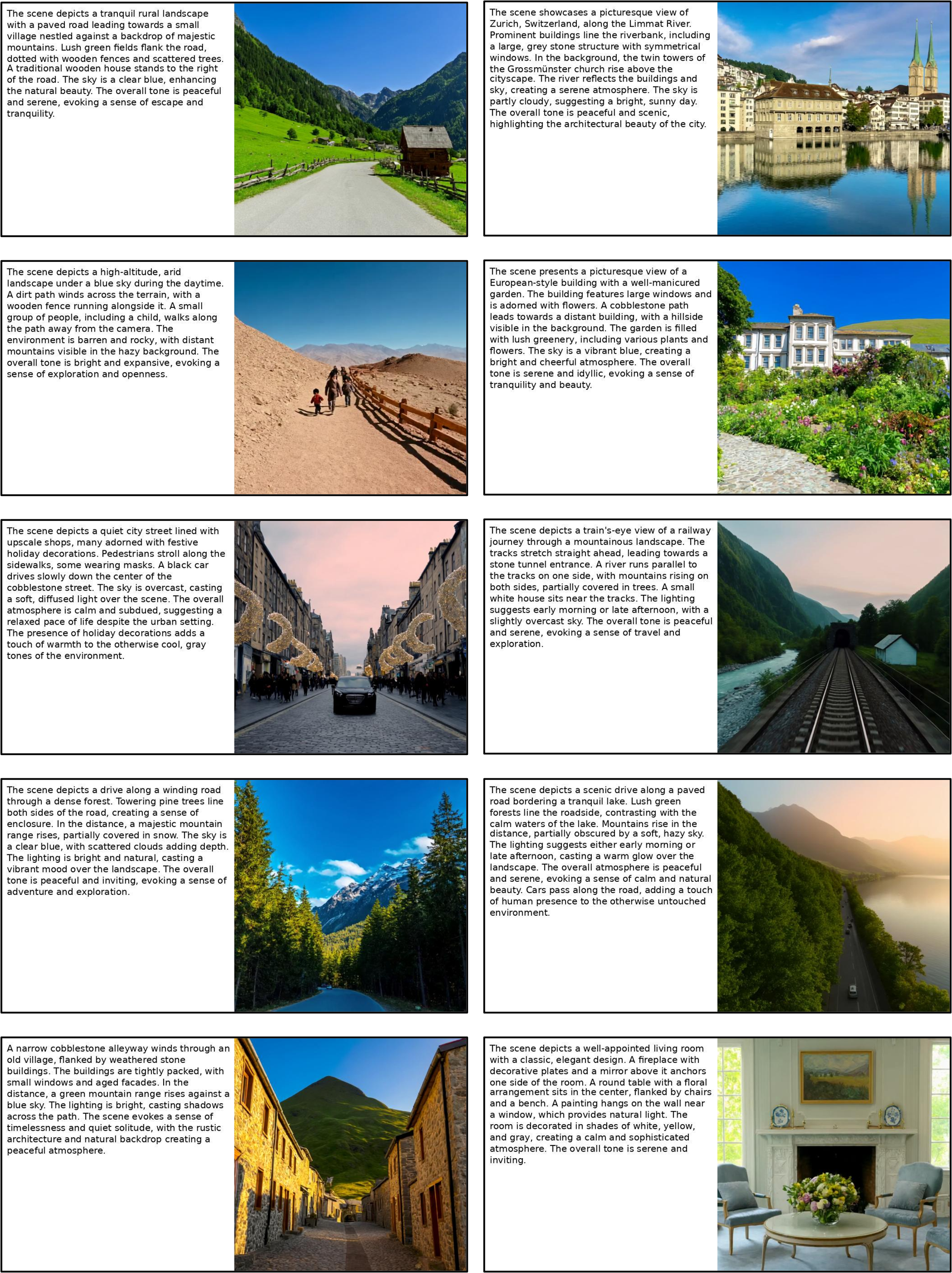}
        
    \end{overpic}
    \caption{Gallery of text-to-image generation results.}
    \label{fig:vis_t2i}
\end{figure*}

To assess the effectiveness of our approach, we provide qualitative results in Fig.~\ref{fig:vis_t2i}. As shown, our model generates images that are not only visually compelling but also semantically faithful to the content encoded in the painted text prompts. The generated samples accurately capture described objects, attributes, and scene structures, demonstrating that visually represented text provides a strong and interpretable conditioning signal. Moreover, the model is robust to variations in layout, font style, and formatting in the painted text, indicating that it learns a consistent visual-language grounding within a unified representation space.

These results collectively show that visualizing text as images is a viable and effective strategy for conditioning image generation, enabling seamless integration of understanding and generation capabilities within a single diffusion transformer

\subsection{Image Understanding}

For image understanding tasks, our model takes an RGB image as input and predicts a painted text image that visually describes semantic outputs. This formulation eliminates the need for text decoders or multimodal fusion modules; instead, all reasoning and prediction occur entirely within the visual latent domain.

\begin{figure*}
    \centering
    \begin{overpic}[width=0.92\linewidth]{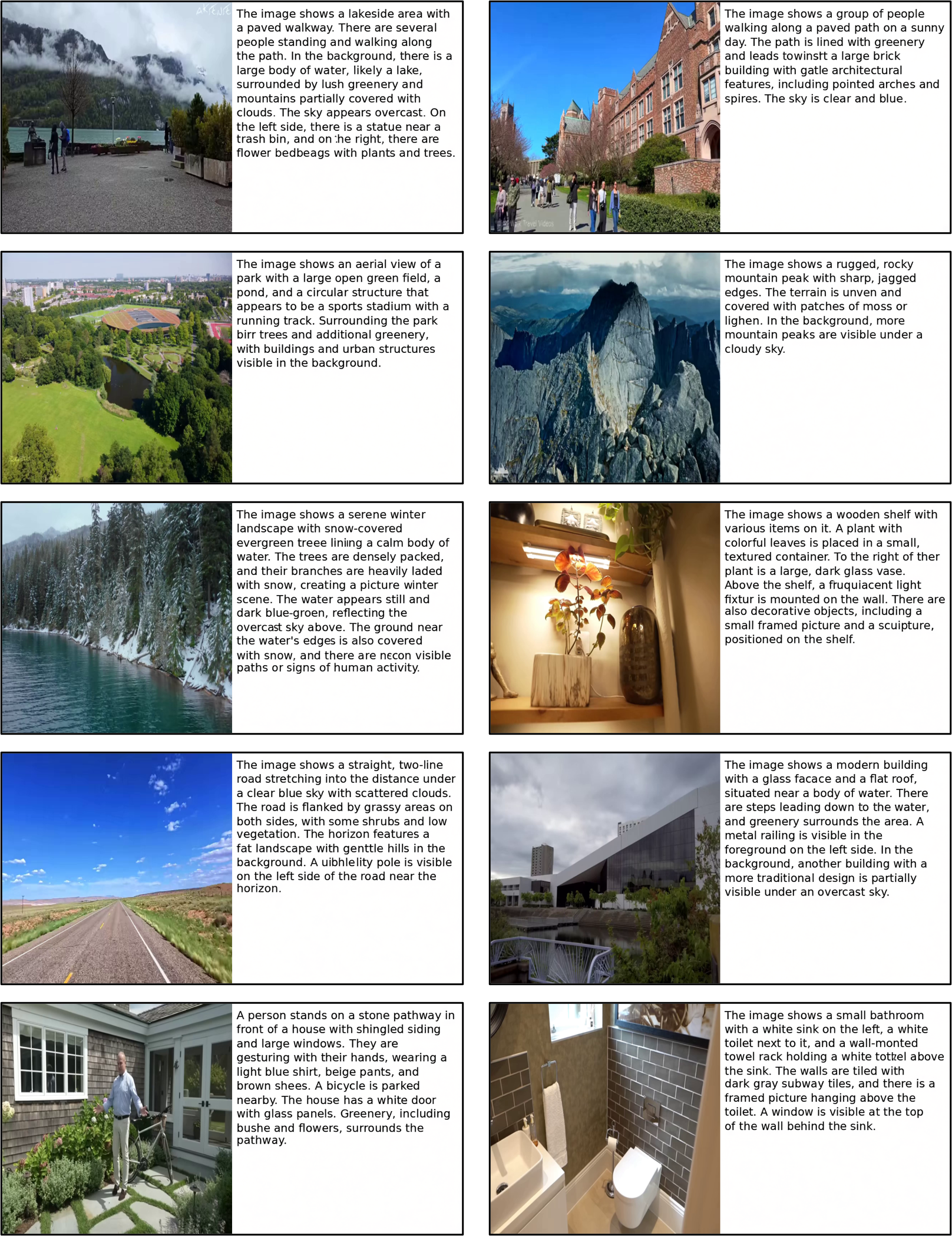}
        
    \end{overpic}
    \caption{Gallery of image-to-textimage generation results.}
    \label{fig:vis_i2t}
\end{figure*}

To demonstrate the effectiveness of this unified representation, we present qualitative results in Fig.~\ref{fig:vis_i2t}. As shown, the model produces text outputs that are both semantically accurate and spatially well-structured. The painted predictions not only reflect correct high-level semantics—such as object descriptions, attributes, and relationships—but also exhibit coherent layout and formatting. In particular, the model autonomously organizes text spatially (e.g., line breaking and alignment), suggesting that the visual diffusion transformer implicitly learns both language semantics and text layout conventions.

These results indicate that our unified visual framework supports robust image understanding, validating the versatility of directly representing textual information in visual form.

\subsection{Analysis}

Thanks to its unified architecture and bidirectional training objective, UniModel naturally supports \textbf{cycle-consistent inference}:
$
\text{RGB} \rightarrow  \text{painted text} 
\rightarrow \text{reconstructed RGB}
$.

\begin{figure}
    \centering
    \includegraphics[width=\linewidth]{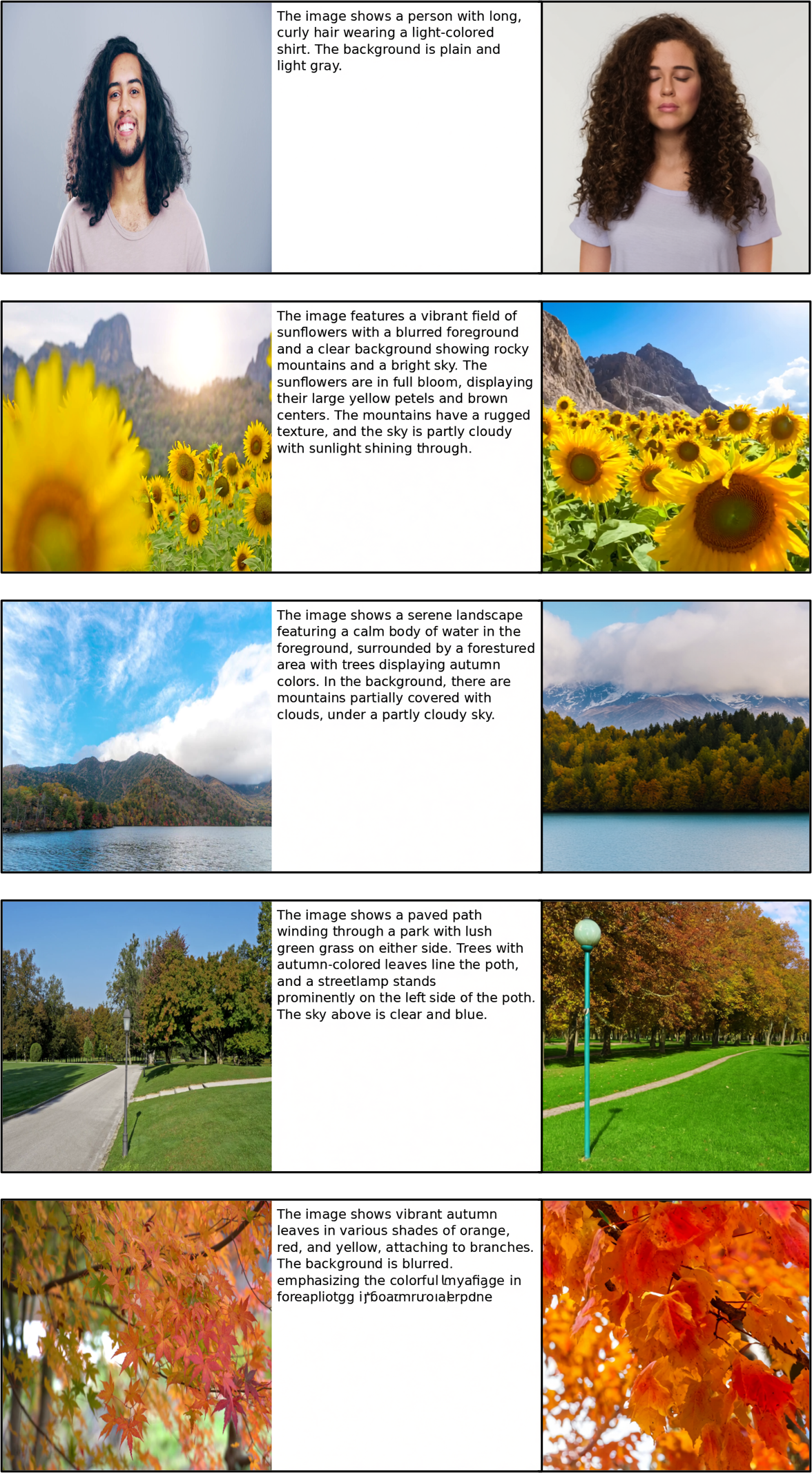}
    \caption{
    Cycle inference results. Starting from an input image, the model generates a painted text description and then reconstructs an image from the painted text. The reconstructed image preserves the key semantics and visual attributes of the original, demonstrating strong bidirectional consistency in our unified visual representation.
    }
    \label{fig:cycle_infer}
\end{figure}

Given an input RGB image $\mathbf{x}_\text{rgb}$, the model first predicts a \emph{painted text image}, which visually describes high-level semantics such as object categories, attributes, and spatial relations. Although this intermediate representation may contain rendering imperfections (e.g., distorted glyphs or incomplete words), it generally remains semantically faithful.
Conditioning the model on $\mathbf{x}_\text{text}$ with $\text{task=generation}$ then produces a reconstructed RGB image $\hat{\mathbf{x}}_\text{rgb}$ that preserves the original semantics while often exhibiting clean compositions and consistent visual structure.
The generative process is resilient to imperfections in the painted-text input, indicating strong semantic grounding and cross-modal consistency within the shared visual space.
Fig. \ref{fig:cycle_infer} shows a gallery of cycle inference results.

These observations suggest that UniModel learns a unified visual representation in which visual and textual representations are unified in visual space—despite the absence of any explicit cycle-consistency loss during training.

\subsection{Limitations}
\begin{figure}
    \centering
    \begin{overpic}[width=0.95\linewidth]{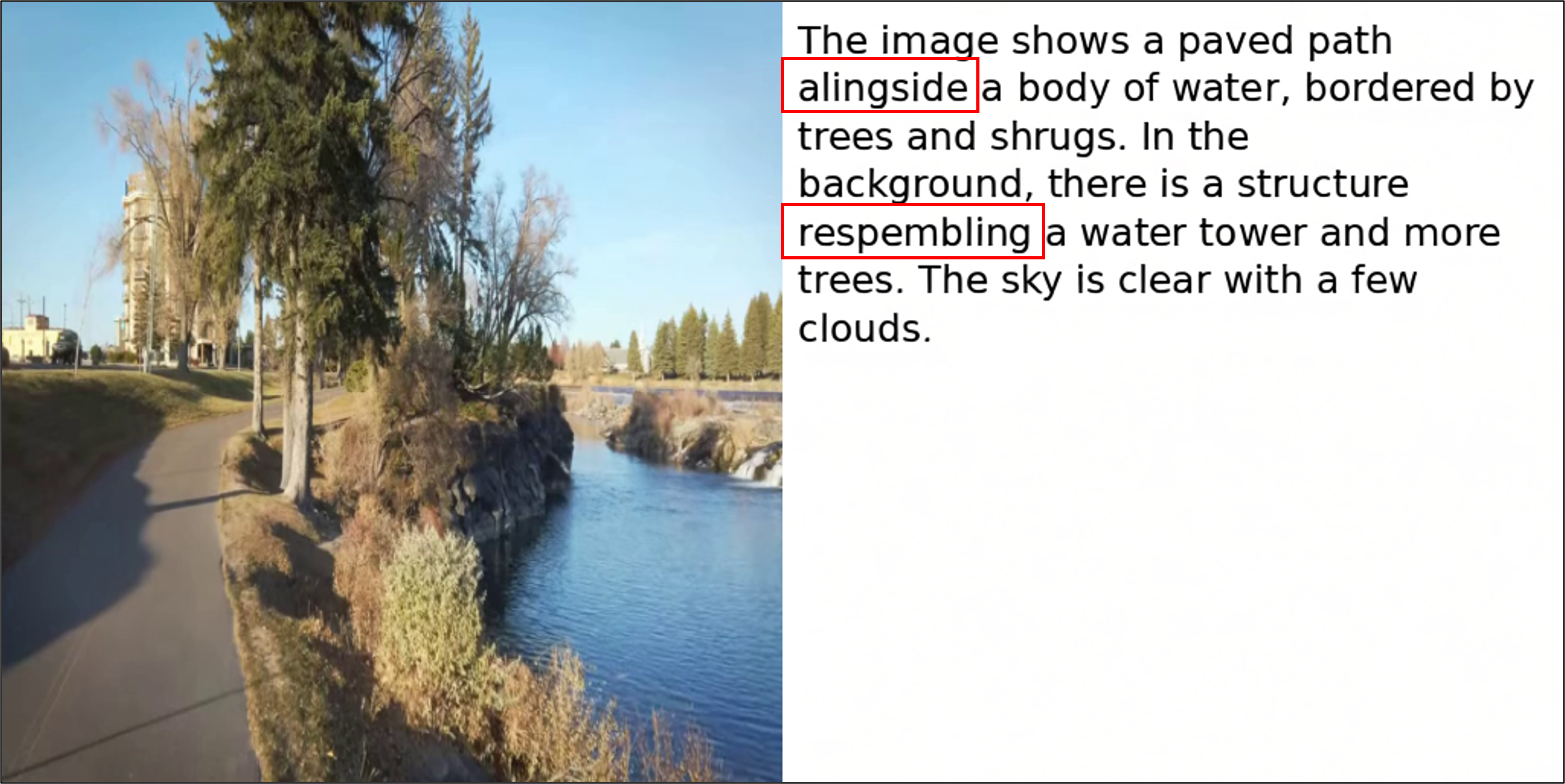}
        
    \end{overpic}
    \caption{Failure case. Although the description is semantically reasonable, the painted text includes incorrect characters and misspelled words, highlighting remaining challenges in accurate word-level generation.}
    \label{fig:limitation}
\end{figure}

While UniModel achieves strong unification across vision understanding and generation, several limitations remain:
\textbf{(1) Training efficiency.} Jointly optimizing for bidirectional tasks (RGB $\leftrightarrow$ text rendering) increases optimization complexity, leading to longer convergence times than single-direction baselines under identical hardware/resources. This stems from the need to balance gradient signals across heterogeneous objectives and stabilize the shared representation space.
\textbf{(2) Text rendering fidelity.} The rendered text images—though semantically meaningful—can exhibit visual artifacts, such as font inconsistencies, glyph distortions (Fig.~\ref{fig:limitation}). While the generative head is robust to mild imperfections (Fig.~\ref{fig:cycle_infer}), severe artifacts may degrade downstream reconstruction quality or mislead fine-grained understanding.

We view these challenges as promising directions for future work: e.g., incorporating perceptual text priors, adaptive curriculum learning, or lightweight task routing to improve efficiency and fidelity.

\section{Conclusion and Future Work}
We introduced UniModel, a Unified Diffusion Transformer that integrates visual understanding and generation within a single, modality-agnostic framework. By representing both natural images and textual content as pixel-based inputs, UniModel removes the need for language tokenization, modality-specific encoders, and cross-modal fusion modules. Combined with bidirectional training, task-aware embeddings, and a rectified-flow objective, this unified design enables a single model to perform image–text and text–image transformations with strong cross-modal alignment and emergent properties such as cycle-consistent inference.

Despite remaining challenges—such as improving text rendering fidelity and further scaling training efficiency—UniModel demonstrates the potential of treating perception and generation as two directions of the same pixel-to-pixel mapping. We believe this work provides a step toward general-purpose visual foundation models that learn, reason, and synthesize entirely within a unified visual space, and we hope it motivates future research into fully end-to-end multimodal systems built on shared pixel-level representations.

We believe UniModel represents a step toward general-purpose visual foundation models that learn, reason, and synthesize entirely within a unified visual space, and we hope this work inspires further research into end-to-end multimodal systems grounded in shared pixel-level representations.

{
    \small
    \bibliographystyle{ieeenat_fullname}
    \bibliography{main}
}


\end{document}